\documentclass[runningheads]{llncs}
\usepackage{float}
\usepackage[table,xcdraw]{xcolor}
\usepackage{amsmath,amssymb,amsfonts}
\usepackage{setspace}
\usepackage{graphicx}
\usepackage{hyperref}

\DeclareMathOperator*{\argmin}{argmin}

\begin{document}
\title{Modelling Agent Policies\\ with Interpretable Imitation Learning\thanks{Supported by an EPSRC/Thales industrial CASE award in autonomous systems.}}
%
%
\author{Tom Bewley\orcidID{0000-0002-5460-0744},
Jonathan Lawry\orcidID{0000-0002-3488-5682},\\ and
Arthur Richards\orcidID{0000-0001-9500-5514}}
\authorrunning{T. Bewley et al.}
%
\institute{University of Bristol, Bristol, UK\\
\email{\{tom.bewley,j.lawry,arthur.richards\}@bristol.ac.uk}}
\maketitle              
\begin{abstract}
As we deploy autonomous agents in safety-critical domains, it becomes important to develop an understanding of their internal mechanisms and representations. We outline an approach to imitation learning for reverse-engineering black box agent policies in MDP environments, yielding simplified, interpretable models in the form of decision trees. As part of this process, we explicitly model and learn agents' latent state representations by selecting from a large space of candidate features constructed from the Markov state. We present initial promising results from an implementation in a multi-agent traffic environment.

\end{abstract}

\section{Introduction}
Data-driven learning is state-of-the-art in many domains of artificial intelligence (AI), but raw statistical performance is secondary to the trust, understanding and safety of humans. For autonomous agents to be deployed at scale in the real world, people with a range of backgrounds and remits must be equipped with robust mental models of their learning and reasoning. However, modern learning algorithms involve complex feedback loops and lack semantic grounding, rendering them black boxes from a human perspective. The field of explainable artificial intelligence (XAI) \cite{samek2017explainable} has emerged in response to this challenge. 

Most work in XAI focusses on developing insight into classification and regression systems trained on static datasets. In this work we consider dynamic problems comprising agents interacting with their environments. We present our approach to interpretable imitation learning (I2L), which aims to model the policy of a black box agent from analysis of its input-output statistics. We call the policy model \textit{interpretable} because it takes the form of a binary decision tree, which is easily decomposed and visualised, and can be used for both factual and counterfactual explanation \cite{guidotti2019factual}. We move beyond most current work in the imitation learning literature by explicitly learning a latent state representation used by the agent as the basis for its decision making. After formalising our approach, we report the initial results of an implementation in a traffic simulator.

\section{I2L Framework} \label{se:framework}

\subsubsection{Preliminaries}
Our I2L approach is applied to agents that operate in Markov Decision Process (MDP) environments. At time $t$, an agent perceives the current Markov state $s_t\in\mathcal{S}$. We then assume that it maps this state into an intermediate representation $x_t=\phi(s_t)$ then selects an action from the discrete space $\mathcal{A}$ via a deterministic policy function $a_t=\pi(x_t)$. The next Markov state $s_{t+1}$ is a function of both $s_t$ and $a_t$. Modelling a state-dependent reward function is not necessary for the present work.

\vspace{-0.2cm}
\subsubsection{Generic Problem Formulation} 
We adopt the perspective of a passive spectator of the MDP with access to its Markov state and the action taken by the agent at each time step. This allows us to observe a history of $N$ states and actions $\mathcal{H}=[(s_1,a_1),...,(s_N,a_N)]$ to use as the basis of learning. Our objective is to imitate the agent's behaviour, that is, to reverse-engineer the mapping $s_t\rightarrow a_t$. This effectively requires approximations of both $\phi$ and $\pi$, denoted by $\phi'$ and $\pi'$ respectively. The need to infer both the policy and the representation on which this policy is based makes this problem a hybrid of imitation learning and state representation learning \cite{lesort2018state}. 

\begin{figure}[H]
\centering
\includegraphics[width=0.85\textwidth]{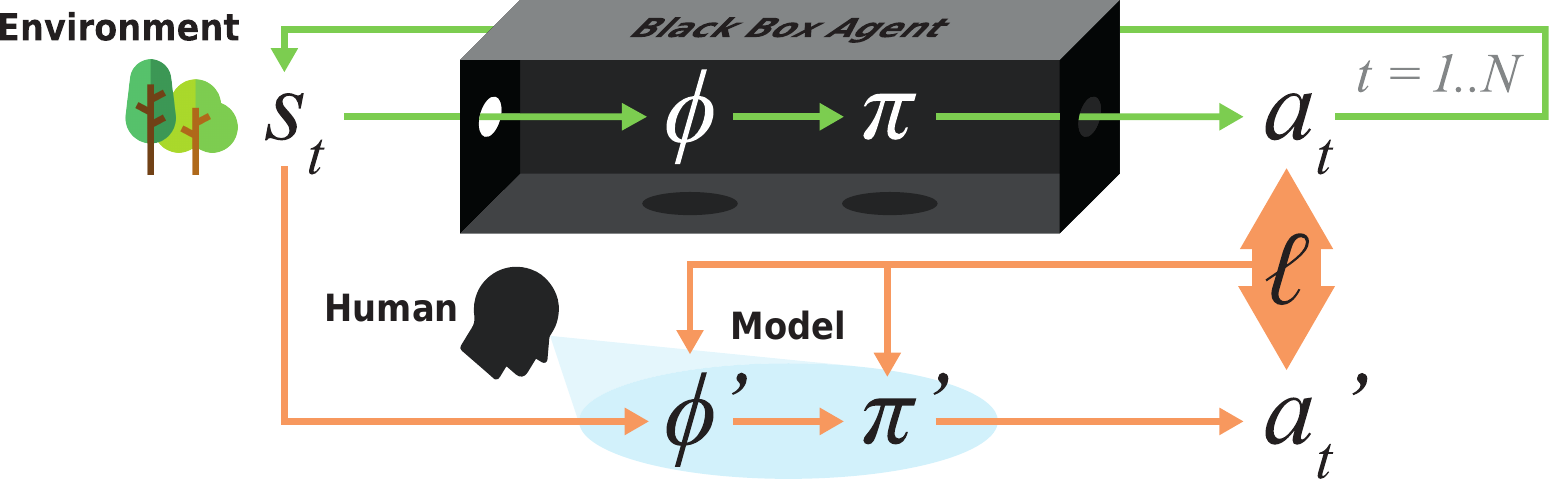}
\caption{Generic I2L problem setup. The objective is to minimise the loss between $a_t$ and $a'_t$ $\forall t\in1..N$, while ensuring $\phi'$ and $\pi'$ are comprehensible under human scrutiny.}
\label{fig:schematic}
\end{figure}

It is essential to constrain the search spaces for $\phi'$ and $\pi'$ ($\Phi$ and $\Pi$ respectively) so that they only contain functions that are human-interpretable, meaning that their operation can be understood and predicted via visualisations, natural-language explanations or brief statements of formal logic. This property must be achieved while minimally sacrificing imitation quality or tractability of the I2L problem. Given the history of state-action pairs $\mathcal{H}$, this problem can be formulated as an optimisation over $\phi'$ and $\pi'$:
\begin{equation}
\label{eq:objective}
\argmin_{\phi'\in\Phi,\pi'\in\Pi}\left[\sum_{t=1}^N\ell\left(\pi'(\phi'(s_t)),a_t\right)\right]\ \ \text{where}\ \ \Phi,\Pi=\textit{``interpretable"}
\end{equation}

\noindent and $\ell:\mathcal{A}\times\mathcal{A}\rightarrow\mathbb{R}_{\geq 0}$ is a pairwise loss function over the discrete action space. The schematic in figure \ref{fig:schematic}\footnote{Icons from users \textit{Freepik} and \textit{Pixel Perfect} at \url{www.flaticon.com}.} outlines the task at hand.

\vspace{-0.4cm}
\newpage
\subsubsection{Space of Representation Functions}
As highlighted above, it is important that $\Phi$ permits human-interpretable state representations. We achieve this by limiting its codomain to vectors of real-valued features, generated from $s_t$ through recursive application of elementary operations from a finite set $\mathbb{F}$. To limit the complexity of $\Phi$, we specify a limit $r$ on the recursion depth. Prior domain knowledge is used to design $\mathbb{F}$, which in turn has a shaping effect on $\phi'$ without requiring the stronger assumption of which precise features to use. Each feature also has a clear interpretation by virtue of its derivation via canonical operations. In our traffic simulator implementation, $\mathbb{F}$ contains operations to extract a vehicle's speed/position, find the nearest vehicle/junction ahead of/behind a position, and subtract one speed/position from another.

\vspace{-0.3cm}
\subsubsection{Space of Policy Functions}
$\Pi$ must be similarly limited to functions that are comprehensible to humans while retaining the representational capacity for high-quality imitation. Following prior work \cite{coppens2019distilling,turnbull2016cloned}, we achieve this by adopting a decision tree structure. The pairwise loss function $\ell$ over the action space $\mathcal{A}$ is used to define an impurity measure for greedy tree induction. For a decision node $n$, let $P_n(a)$ be the proportion of data instances at that node with action value $a$. We use the following measure of node impurity:

\begin{equation}
\label{eq:impurity}
I(n)=\sum_{a\in\mathcal{A}}\sum_{a'\in\mathcal{A}}P_n(a)\cdot P_n(a')\cdot \ell(a,a')
\end{equation}

\noindent The popular Gini impurity is a special case of this measure, recovered by defining $\ell$ such that $\ell(a,a')=0$ if $a=a'$, and $\ell(a,a')=1$ otherwise. 

\vspace{-0.3cm}
\subsubsection{Learning Procedure} 
\label{ssec:procedure} In general, the joint inference of two unknown functions (in this case $\phi'$ and $\pi'$) can be very challenging, but our constraints on $\Phi$ and $\Pi$ allow us to approximately solve equation \ref{eq:objective} through a sequential procedure:

\begin{enumerate}
\item Apply domain knowledge to specify the feature-generating operations $\mathbb{F}$ and recursion depth $r$. Denote the representation of all valid features $\phi_{\text{all}}$. 
\item Iterating through each state $s_t$ in the observed state-action history $\mathcal{H}$, apply $\phi_{\text{all}}$ to generate a vector of numerical feature values. Store this alongside the corresponding action $a_t$ in a training dataset.
\item Define a pairwise action loss function $\ell$ and deploy a slightly-modified version of the CART tree induction algorithm \cite{CART} to greedily minimise the associated impurity measure (equation \ref{eq:impurity}) on the training set. Let the induction process continue until every leaf node is pure, yielding a large, overfitted tree $\mathcal{T}_{0}$.
\item Prune the tree back using minimal cost complexity pruning (MCCP) \cite{CART}, whose output is a tree sequence $[\mathcal{T}_{0},\mathcal{T}_{1},\mathcal{T}_{2},...]$, each a subtree of the last, representing a progressive reduction of $\mathcal{T}_{0}$ down to its root.
\item Pick a tree from this sequence to use as the policy model $\pi'$, and define $\phi'$ as the subset of features from $\phi_{\text{all}}$ used at least once in that tree.
\end{enumerate}
Having a sequence of options for the tree model and associated representation allows us to manage a tradeoff between accuracy on one end, and interpretability (through simplicity) on the other. In the following implementation, we explore the tradeoff by selecting several trees from across the pruning sequence. 

\section{Implementation with a Traffic Simulator} \label{se:implementation}
We implement I2L in a traffic simulator, in which multiple vehicles follow a common policy to navigate a track while avoiding collisions. Five track topologies are shown in figure \ref{fig:quality} (left). Coloured rectangles are vehicles and red circles are junctions. Since the policy is homogeneous across the population we can analyse the behaviour of all vehicles equally to learn $\phi'$ and $\pi'$.

\vspace{-0.4cm}
\subsubsection{Target Policies}
Instead of learned policies, we use two hand-coded controllers as the targets of I2L. From the perspective of learning these policies remain opaque. For both policies, $\mathcal{A}$ contains five discrete acceleration levels, which determine the vehicle's speed for the next timestep within the limits $[0,v_\text{max}]$. 
\vspace{-0.05cm}
\begin{itemize}
\item Fully-imitable policy ($\pi_{\text{F}}$): a rule set based on six features including the vehicle's speed, distance to the next junction, and separations and speeds relative to other agents approaching that junction. This policy is itself written as a decision tree with $39$ leaves, hence can be modelled exactly in principle.
\item Partially-imitable policy ($\pi_{\text{P}}$): retains some logic from $\pi_{\text{F}}$, but considers more nearby vehicles and incorporates a proportional feedback controller. These changes cannot be modelled exactly by a finite decision tree.
\end{itemize}

\vspace{-0.55cm}
\subsubsection{Representation}
We use a set of eight operations $\mathbb{F}$ which allows the six-feature representation used by $\pi_{\text{F}}$ (denoted $\phi_{\text{F}}$) to be generated. These are:
\vspace{-0.05cm}
\begin{itemize}
\item $\operatorname{pos}:i\rightarrow p$ or $j\rightarrow p$. Get the position of vehicle $i$ or junction $j$.
\item $\operatorname{speed}:i\rightarrow v$. Get the speed of vehicle $i$.
\item $\operatorname{fj}:i\rightarrow j$. Find the next junction in front of vehicle $i$.
\item $\operatorname{fa}:i\rightarrow i'$ or $j\rightarrow i'$. Find the next vehicle in front of vehicle $i$ or junction $j$.
\item $\operatorname{ba}:i\rightarrow i'$ or $j\rightarrow i'$. Find the next vehicle behind vehicle $i$ or junction $j$.
\item $\operatorname{twin}:j\rightarrow j'$. Flip between the two track positions comprising a junction.
\item $\operatorname{sep}:(p_1,p_2)\rightarrow s$. Compute the separation between two positions.
\item $\operatorname{sub}:(v_1,v_2)\rightarrow \Delta$ or $(s_1,s_2)\rightarrow \Delta$.  Subtract two speeds or separations.
\end{itemize}
\vspace{-0.05cm}
\noindent With $r=6$, $\phi_\text{all}$ has $308$ features, including all six in $\phi_\text{F}$. The vast majority are irrelevant for imitating the target policies, so we face a feature selection problem.

\vspace{-0.4cm}
\subsubsection{Training}
\begin{figure}[]
\centerline{\includegraphics[width=0.75\textwidth]{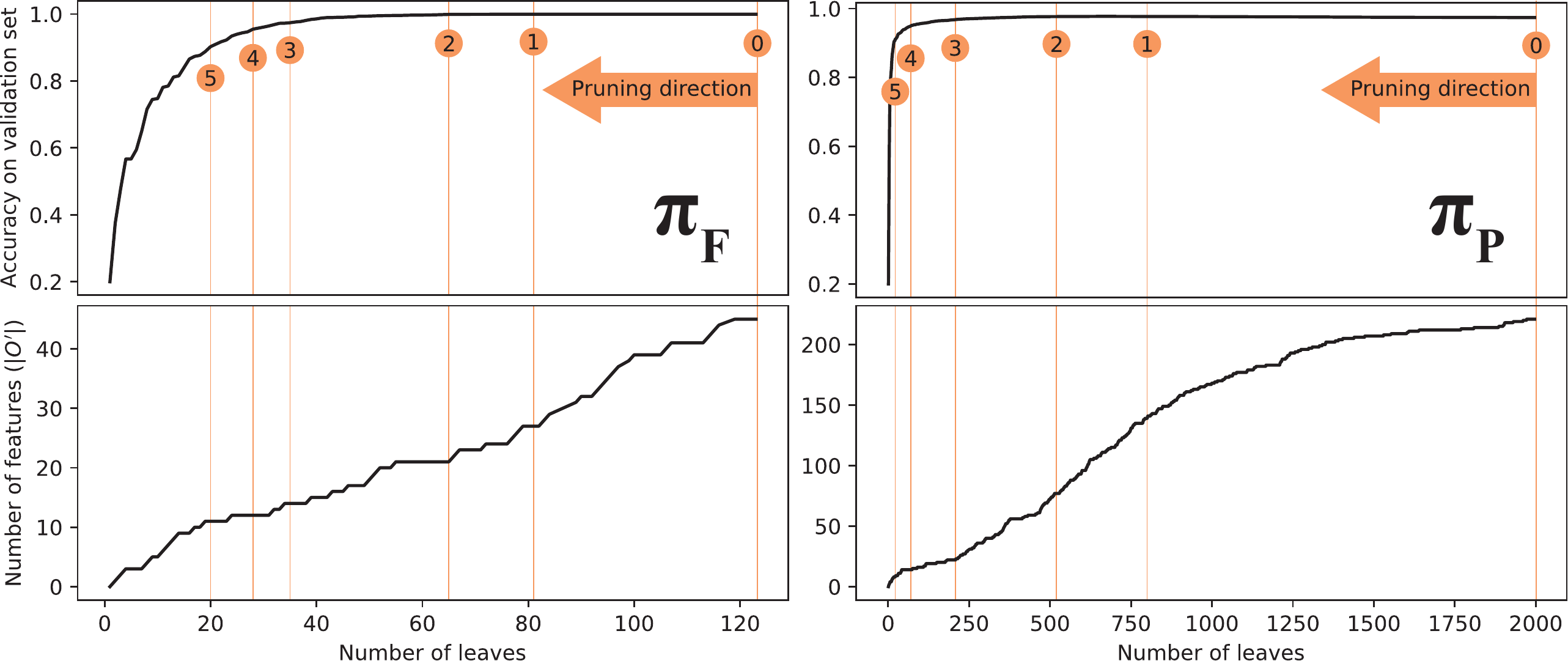}}
\vspace{-0.2cm}
\caption{Validation accuracy and used feature count across the pruning sequences for each target policy. \textit{*Note the very different horizontal axis scales*}; the unpruned tree for $\pi_\text{P}$ is much larger. Numbered vertical lines indicate trees chosen for evaluation.}
\label{fig:pruning}
\end{figure}

We run simulations with $11$ vehicles on all five topologies in figure \ref{fig:quality} (left). The recorded history $\mathcal{H}$ is converted into a training dataset by applying $\phi_{\text{all}}$ for each vehicle. After rebalancing action classes, our final dataset has a length of $125000$. For tree induction, we use the simple loss function $\ell(a,a')=|a-a'|$. For each tree in the sequence output by MCCP, we measure the predictive accuracy on a validation dataset, and consider the number of leaf nodes and features used as heuristics for interpretability. Figure \ref{fig:pruning} plots these values across the sequence for both targets. We select five pruning levels (in addition to $\mathcal{T}_0$) for evaluation.

\vspace{-0.4cm}
\subsubsection{Baselines}
We also train one tree using only the six features in $\phi_{\text{F}}$, and another with an alternative representation $\phi_{\text{na\"{i}ve}}$, indicative of what might be deemed useful without domain knowledge (radius, angle, normal velocity and heading of nearby agents in egocentric coordinates). Two further baselines use training data from only one track topology (either the smallest topology A or the largest D). This tests the generalisability of single-topology learning. For all baselines, we use the pruned tree with the highest validation accuracy.

\newpage

\vspace{-1.05cm}
\section{Results and Discussion} \label{se:results}
\begin{figure}[]
\centerline{\includegraphics[width=\textwidth]{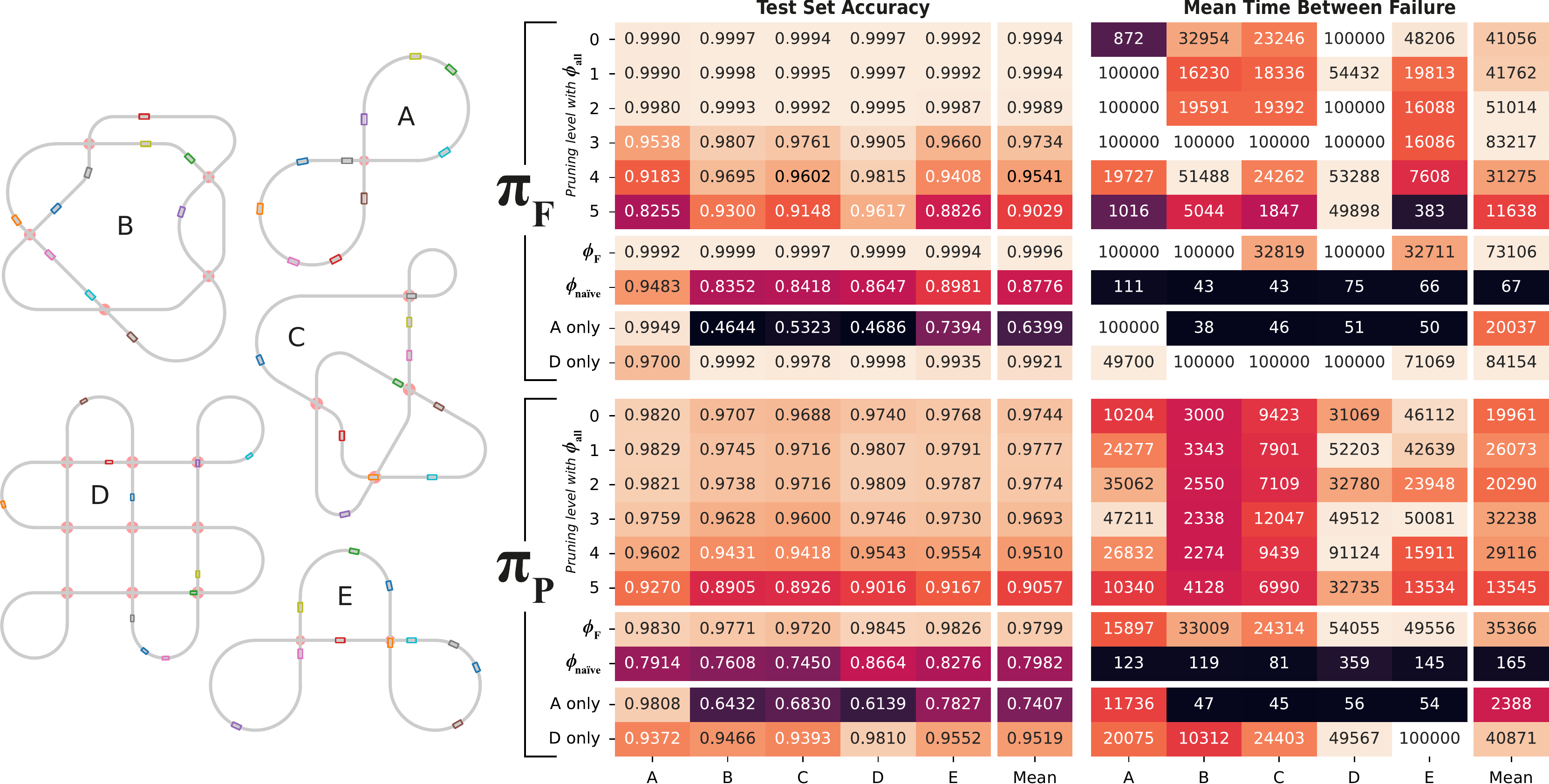}}
\vspace{-0.2cm}
\caption{(Left): The five test topologies used. (Right): Results for I2L quality metrics.}
\label{fig:quality}
\end{figure}

The heat maps in figure \ref{fig:quality} (right) contain results for two metrics of imitation quality for both $\pi_\text{F}$ and $\pi_\text{P}$. Each row corresponds to a pruned tree or baseline, and each column indicates the track topology used for testing.  

\vspace{-0.4cm}

\vspace{-1.1cm}
\subsubsection{Accuracy}
This metric is predictive accuracy on a held-out test set. For both $\pi_\text{F}$ and $\pi_\text{P}$, mean accuracy exceeds $90\%$ for even the smallest prune levels and $95\%$ for the second-smallest. The fact that accuracy for $\pi_\text{P}$ is bounded at around $97.5\%$ reflects the fact that this policy is not a decision tree so cannot be perfectly imitated by one. As expected, providing $\phi_{\text{F}}$ upfront (thereby removing the representation learning requirement) yields somewhat better accuracy, but it is promising to see that down to prune level $2$, performance differs by under $0.25\%$ for both targets. Lacking information about junctions, the tree using $\phi_{\text{na\"{i}ve}}$ is unable to obtain the same levels of accuracy, demonstrating the importance of choosing the correct representation for imitation. The single-topology training results show that trees generalise well from the large topology D, but poorly from the small topology A. This suggests the latter contains insufficient variety of vehicle arrangements to capture all important aspects of the target policies.

\vspace{-0.4cm}
\subsubsection{Failure}
Here we deploy the models as control policies in the environment and measure the mean time between failures (either collision or `stalls', when traffic flow grinds to a halt) over $100$ episodes of up to $1000$ timesteps. A value of $100000$ indicates zero failures occurred. While results broadly correlate with accuracy, this metric shows a more marked aggregate distinction between $\pi_\text{F}$ and $\pi_\text{P}$, and between different test topologies. Nonetheless, there is minimal degradation with pruning down to level $4$, with a constant average of just $1$-$2$ failures for  $\pi_\text{F}$, and $3$-$4$ for  $\pi_\text{P}$. In fact, it appears that intermediate pruning levels fail less often than the largest trees. While the reason for this is not immediately clear, it may be that having fewer leaves yields less frequent changes of acceleration and smoother motion. Providing $\phi_{\text{F}}$ confers no significant benefit over $\phi_{\text{all}}$, while $\phi_{\text{na\"{i}ve}}$ and single-topology training on A are utterly unable to perform/generalise well.

\vspace{-0.2cm}
\section{Conclusion}
We have introduced our approach to interpretable imitation learning for black box agent control policies that use intermediate low-dimensional state representations. Our models take the form of decision trees, which select from large vectors of candidate features generated from the Markov state using a set of basic operations. The accuracy-interpretability tradeoff is managed by post-pruning. 

Our initial implementation has shown that trees trained by I2L exhibit high predictive accuracy with respect to two hand-coded control policies, and are able to avoid failure conditions for extended periods, even when heavily pruned. It has also highlighted that using a plausible-but-incorrect state representation places a severe limitation on imitation quality, and that learning from data that do not capture the full variation of the environment leads to poor generalisation.

In ongoing follow-up work, we are exploring how our decision tree models can be used to interpret and explain their target policies, and are also implementing I2L with a truly black box policy trained by reinforcement learning.

\vspace{-0.15cm}
{\setstretch{0.6}
\bibliographystyle{splncs04}
\bibliography{Bibliography}{}
}
\end{document}